# Kernaug V1.0 Alpha |

*The type of tasks that falls in the 4 D's of Robotisation: Dull, Dirty, Dangerous and Dear, task taken over by robots*

Aymen Rachdi
Embedded System Developer & Roboticist
Kernel Augmentation
contact@aymen.rachdi.xyz

Amine Kammoun
Full Stack developer
Kernel Augmentation
Aminekammoun55@gmail.com

Fedi Zrelli
Hardware Engineer
Kernel Augmentation
fedizrelli3@gmail.com

**Abstract—** For a robot to be perfect and enter the everyday life of humans, like computers did, it needs to move from special-purpose robots to general-purpose. So, the idea of modularity is considered in this project. Thus, any type of task that falls in the 4 D's of Robotisation: Dull, Dirty, Dangerous and Dear can be achieved by adding a module to the robot.

## I. Introduction

Kernaug is a modular robot, the team working on this project is trying to implement modularity in every part of the robots' core.

The architecture of such robots is sophisticated and full of different options that we can choose from . So the group of developers, roboticists, Embedded System Developers and cyber security experts must pay attention to a lot of caveats in the design.

The robot's modules are themselves modular. We think a lot about the freedom, privacy and security of the user so we try to give various options while maintaining safe, secure and reliable robots.

These types of robots may be used on a daily basis in hospitals and maybe used in response to a crisis like a pandemic. In these types of situations we must provide **Affordability** [1]**:** where any establishment could get this robot for a cheaper price.

**Safety** [2]**:** we need to provide safety as it's the main goal and stop the robot from functioning when it poses risks.

**Manufacturing speed** [3]**:** We need a cheap fast strategy to produce robot parts **[Research in progress]**

**Size and shape complexity** [4]**:** In relation to the price [1] and production speed [3] we must keep the robot's shape minimalistic and elegant.

**Scalability** [5]: As we migrate from ROS1 TO ROS2 we will acquire more features built in the new robot operating system. It will enable us to use docker and kubernetes to scale on demand and for blazing fast deployment of new robots.

**Security** [6]**:** Robots are, for most cases, built insecure and fully unprotected by design. The complexity of a robotic system makes the vendors and robotic companies unable to cover the complete threat landscape.

Cybersecurity in robotics is tricky to implement but essential as no robotic system is safe[2] without security.

## II. Main Features

- *Camera streaming:*
  Continuous camera streaming that allows the user to live a moving experience during the night exploration of a building for remote monitoring.

- *Intelligent obstacle avoidance:*
  Equipped with a new 360-degree omnidirectional obstacle avoidance technology, the robot can perfectly perceive dynamic and suspended obstacles inside and react instantly to automatically bypass obstacles.

- **Autonomous wireless charging systems:**
  Stand-alone wireless charging system with the ability to purchase more than one charging station for larger spaces.

- *Assisted navigation:*
  *The robot is equipped with a joystick implemented in the web application to guide the robot manually.*

- *Simple and clear user interface:*
  *The software side is a progressive web application (pwa) that allows the user to control the robot and perform many functions.*

- *Modules to plug and play*
  This robot is modular; you can add a variety of additional features by plugging new modules into it.

- *Robot's Security and safety feature*
  We provide state of the art security and beyond military grade encryption with performance in mind.

### III. TECHNICAL SPECIFICATIONS

| | |
|---|---|
| *Size:* | *400*400*400 mm* |
| *Network interface* | *WIFI/4G* |
| *App* | *Main robot web app with Add-ons provided to the modules* |
| *Power rating* | *150W* |
| *Lifetime* | *≈20000h* |
| *Work environment* | *Indoor environment, flat and smooth floor* |
| *Operating temperature* | 0 - 45℃ |
| *Maximum angle of climb* | ≤ 5° |
| *Maximum speed* | 0-0.8m/s |
| *Battery capacity* | DC 12V 30Ah expandable to 50Ah |
| *Battery life* | ≈4h |
| *Charging mode* | Autonomous / manual |
| *Charging time* | 4h |
| *Sensor* | Lidar, artificial vision, depth vision, |
| *Minimum passage width* | ≥60cm |

### IV. ELECTRICAL DESIGN

This document presents the electronic and electrical design of the " Kernaug " Robotic Project.
Specification of the hardware used in the "Kernaug" robot:

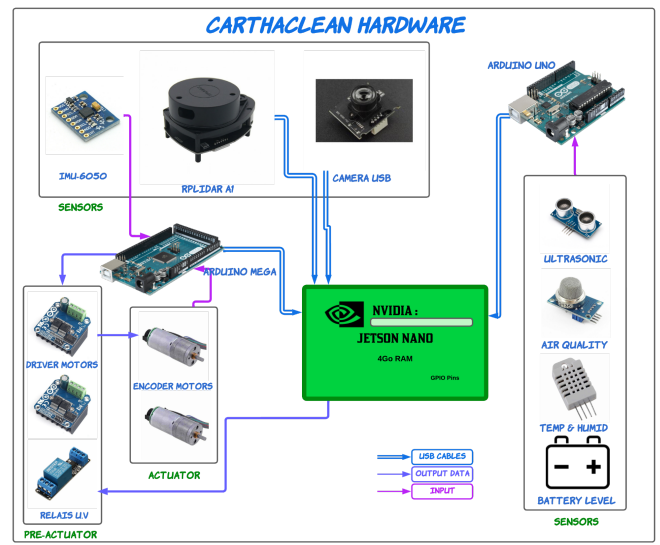

**FIG. 1**  Electrical Design

## Control Unit

- **NVIDIA® Jetson Nano:** Micro-processing Unit for heavy loads, all in an easy-to-use platform that consumes only 5 watts of power.
- **ATmega2560:** microcontroller board based on the ATmega2560 for blazing fast control.

## The sensors

- **RPLIDAR A1:** used for mapping, localization and modeling of objects and environments.
- **Ultrasonic sensor:** obstacle detection sensor based on ultrasonic waves used for obstacle detection in the robot's blindspots.
- **3-axis Accelerometer:** A 3-axis accelerometer measures the accelerations that occur with respect to the 3 Cartesian coordinate axes. In other words, it can measure changes in the velocity of a point. In addition, it can also detect the direction and speed of a moving object, taking into account the dynamic acceleration…
- 

## The pre-actuator and the actuator

- **Driver Motor:** The BTS7960 is a fully integrated high current half-bridge for motor drive applications. With an operating voltage of 24Vdc and a peak current of 43A, it has a PWM capability of up to 25 kHz combined with active freewheeling. The BTS7960 is a fully integrated high current H-bridge module for motor drive applications.
- **Gearmotor with Encoder:** This JGA25-371 DC gearmotor features an integrated encoder with 12 counts per revolution resolution, ensuring precise control of motor speed.

## Power Supply and Regulation

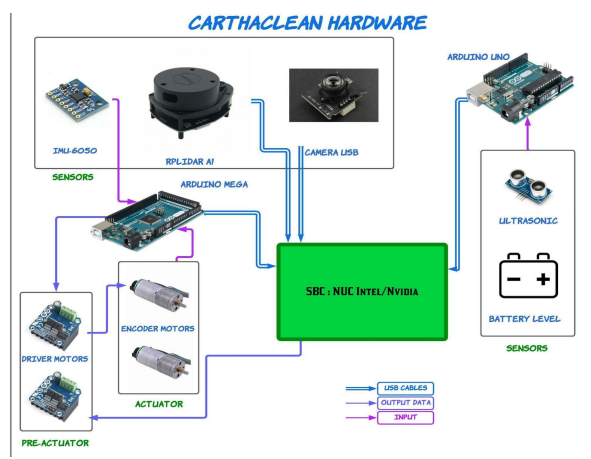

**Fig. 2** Power Supply

- **Battery:** we made a battery according to our needs, based on a set of rechargeable lithium batteries, each battery block provided 11.1V / 8000mAh and dimension.
- **Booster converter:**

  with an input voltage of DC 6-35V this booster can output 6-55V DC. The Conversion efficiency is 97.6% and Maximum output current: 7A.

  **LM2596 regulator:** it can handle 3A but for long term use an amperage of 2-2.5A is recommended, for input voltage from 3.2V to 40V and output voltage from 1.25V to 35V...

## V. Mechanical Design

- **Gearmotor with Encoder:**

**3D design**

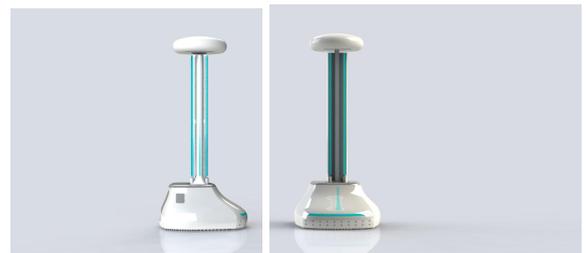

## VI. Software

We used Ubuntu 20.0.4 as our operating system as it's the stable long term support available system at the time of writing of this paper and we used ROS_DISTRO:NOETIC as our robotic framework or metaoperting system.

### 1. Bringup launch file

Preparing the launch file for a robot is tricky because all other steps rely on it. The robot Bringup launch file contains all different nodes and parameters to start and set up all sensors.

Nodes:

- **rosserial**: rosserial is a protocol for wrapping standard ROS serialized messages and multiplexing

### 2. Mapping (SLAM)

The SLAM (Simultaneous Localization and Mapping) is a technique to draw a map by estimating current location in an arbitrary space.

Nodes:

- Gmapping

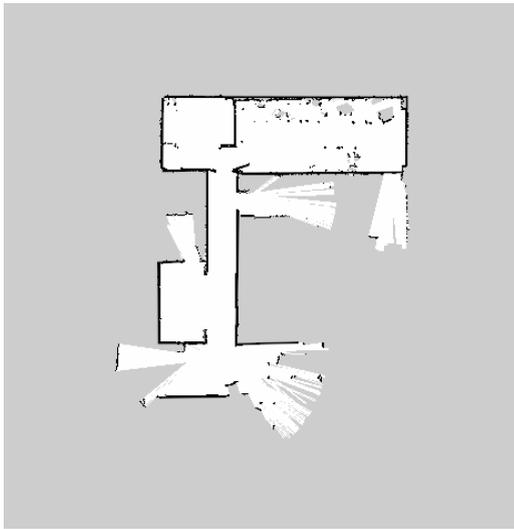

**FIG. 4** Map created moving the robot

### 3. Navigation Stack

In this part it's very important to know every part and its physical characteristics. Like the robot height and width, distance from wheel to wheel, distance from Lidar sensor to base and a lot of other parameters we should set for the robot to function. Any information set inaccurately could render the robot nonfunctional or behave unexpectedly.

A 2D navigation stack that takes in information from odometry, sensor streams, and a goal pose and outputs safe velocity commands that are sent to a mobile base.

*amcl* : amcl is a probabilistic localization system for a robot moving in 2D. It implements the adaptive (or KLD-sampling) Monte Carlo localization approach (as described by Dieter Fox), which uses a particle filter to track the pose of a robot against a known map. amcl takes in a laser-based map, laser scans, and transform messages, and outputs pose estimates. On startup, amcl initializes its particle filter according to the parameters provided. Note that, because of the defaults, if no parameters are set, the initial filter state will be a moderately sized particle cloud centered about (0,0,0).

*base_local_planner* :This package provides implementations of the Trajectory Rollout and Dynamic Window approaches to local robot navigation on a plane. Given a plan to follow and a costmap, the controller produces velocity commands to send to a mobile base. This package supports both holonomic and non-holonomic robots, any robot footprint that can be represented as a convex polygon or circle, and exposes its configuration as ROS parameters that can be set in a launch file. This package's ROS wrapper adheres to the BaseLocalPlanner interface specified in the nav_core package.

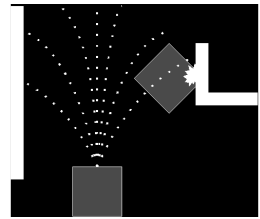

**FIG. 5** Dynamic Window Approach (DWA) algorithms

The basic idea of both the Trajectory Rollout and Dynamic Window Approach (DWA) algorithms is as follows:

- Discretely sample in the robot's control space (dx,dy,dtheta)
- For each sampled velocity, perform forward simulation from the robot's current state to predict what would happen if the sampled velocity were applied for some (short) period of time.
- Evaluate (score) each trajectory resulting from the forward simulation, using a metric that incorporates characteristics such as: proximity to obstacles, proximity to the goal, proximity to the global path, and speed. Discard illegal trajectories (those that collide with obstacles).
- Pick the highest-scoring trajectory and send the associated velocity to the mobile base.
- Rinse and repeat.

DWA differs from Trajectory Rollout in how the robot's control space is sampled. Trajectory Rollout samples from the set of achievable velocities over the entire forward simulation period given the acceleration limits of the robot, while DWA samples from the set of achievable velocities for just one simulation step given the acceleration limits of the robot. This means that DWA is a more efficient algorithm because it samples a smaller space, but may be outperformed by Trajectory Rollout for robots with low acceleration limits because DWA does not forward simulate constant accelerations. In practice however, we find DWA and Trajectory Rollout to perform comparably in all our tests and recommend use of DWA for its efficiency gains.

costmap_2d :This package provides an implementation of a 2D costmap that takes in sensor data from the world, builds a 2D or 3D occupancy grid of the data (depending on whether a voxel based implementation is used), and inflates costs in a 2D costmap based on the occupancy grid and a user specified inflation radius. This package also provides support for map_server based initialization of a costmap, rolling window based costmaps, and parameter based subscription to and configuration of sensor topics.

dwa_local_planner :This package provides an implementation of the Dynamic Window Approach to local robot navigation on a plane. Given a global plan to follow and a costmap, the local planner produces velocity commands to send to a mobile base. This package supports any robot who's footprint can be represented as a convex polygon or circle, and exposes its configuration as ROS parameters that can be set in a launch file. The parameters for this planner are also dynamically reconfigurable. This package's ROS wrapper adheres to the BaseLocalPlanner interface specified in the nav_core package.

global_planner :This package provides an implementation of a fast, interpolated global planner for navigation.

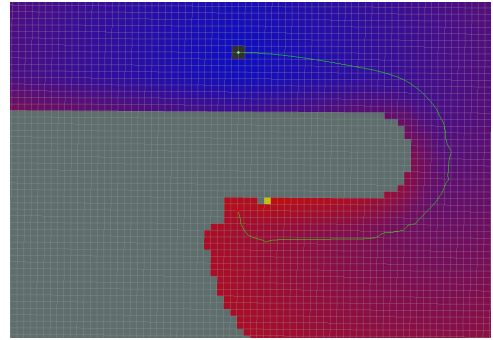

**Fig. 6** global planner

map_server :map_server provides the map_server ROS Node, which offers map data as a ROS Service. It also provides the map_saver command-line utility, which allows dynamically generated maps to be saved to file.

move_base :The move_base package provides an implementation of an action

The actionlib stack provides a standardized interface for interfacing with preemptable tasks. Examples of this include moving the base to a target location, performing a laser scan and returning the resulting point cloud, detecting the handle of a door, etc.

given a goal in the world, will attempt to reach it with a mobile base. The move_base node links together a global and local planner to accomplish its global navigation task.

- Parameters

costmap_common_params:

The navigation stack uses costmaps to store information about obstacles in the world. In order to do this properly, we'll need to point the costmaps at the sensor topics they should listen to for updates

Local_costmap_params:

Global_costmap_params:

*Base_local_planner_default_params:*

*Move_base_params:*

### 4. Additional Sensors

⇋ Ultrasonic:

We added the ultrasonic sensor to detect obstacles at the bottom of the robot. We fused its data with the data coming from the lidar to use it in the navigation.
*range_sensor_layer*

⇋ camera:

The camera is for docking and streaming.

⇋ Battery sensor

Provide feedback battery level

## VII. COMMUNICATION ARCHITECTURE AND SECURITY

A group of security researcher launched a worldwide scan for ros enabled robots and they found a lot of them[1]

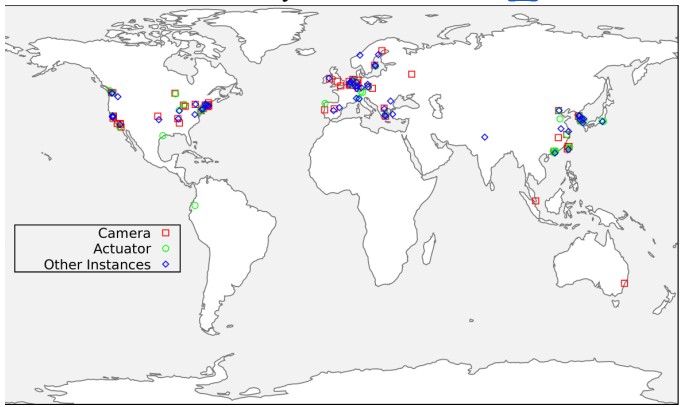

Most of these machines are not resilient to security attacks. Manufacturers' concerns, as well as existing standards, focus mainly on safety. Security is not being considered as a primary relevant matter.

When checking the RHM(Robot Hacking Manual), we found a huge numbers of flaws and vulnerabilities even in ROS2
So the theory of "No system is safe" is confirmed.

*Network segregation and segmentation: the server and the robot are on different subnetwork using openvpn*

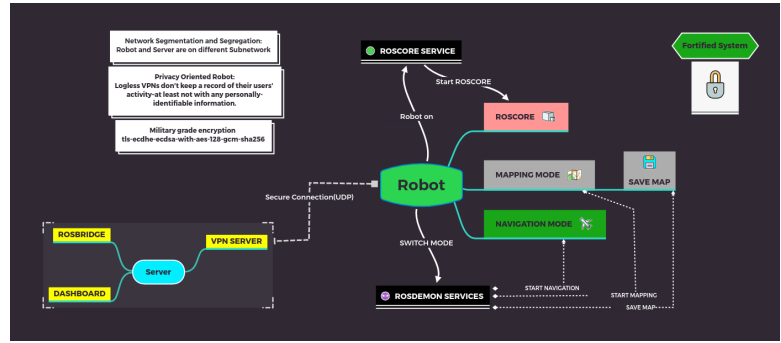

⇋ **OPEN VPN:** A virtual private network to link the robot to the server with ECDH encryption.

The robot's security can never be compromised since only two machines are able to be connected to the server with a private unique key : The robot and the technical support team.

⇋ **FIREWALL:** Firewall set to protect the server from DDOS attacks and prevent all unauthorized access.
⇋ **Logless Server:** With Verb0 enabled the server only saves critical logs, all flowing data will not be recorded including Camera streams. Data will be saved if the user explicitly asks for it.
⇋ **Access notifier**: The user will be notified when someone gains access to the robot.
⇋ **Tamper proof robot:LUKS(Linux Unified Key Setup):** With physical access to the robot the internal software can't be compromised.
⇋ **Military grade encryption: tls-ecdhe-ecdsa-with-aes-128-gcm-sha256:**
⇋ **Robot server:** The installation of the robot includes cloud solution and/or local server installation by the technical team.
⇋ **Encapsulated Desktop application:** can be included for extreme security solutions.

- **Diag.key:** A unique key generated by the Technical Team for maintenance purposes
- **Hardware Diagnosis and maintenance:** Like any electronic system, this robot requires maintenance from time to time so we prepared a guide for the user to do it by himself. This guide will be included with robots. All diagnosis and maintenance procedures are done remotely if, after a period of time, a hardware problem appears we can send part replacement parts and procedure to install them safely and correctly. Clear instructions will be provided on how to change any damaged part so that the user or a non trained technician can follow and replace it.
- **Hardware modularity:** hardware modularity is so hard to implement but our team is working on implementing it even partially. By then, the robot will accept a variety of sensors to plug and play like a computer mouse or keyboard, the robot will recognize them automatically and start using them.

## VIII. MAIN APPLICATION

Intuitive Secure app :

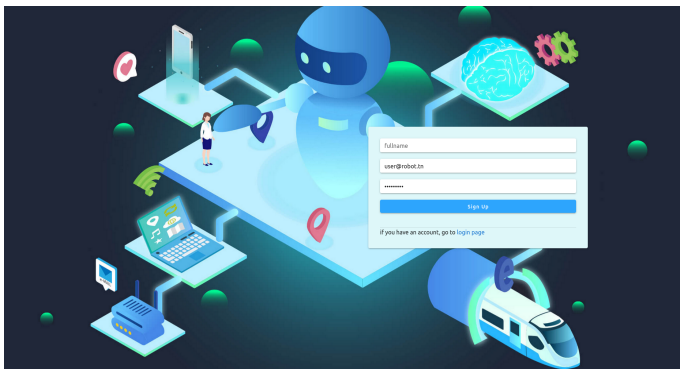

**FIG. 7** Create account

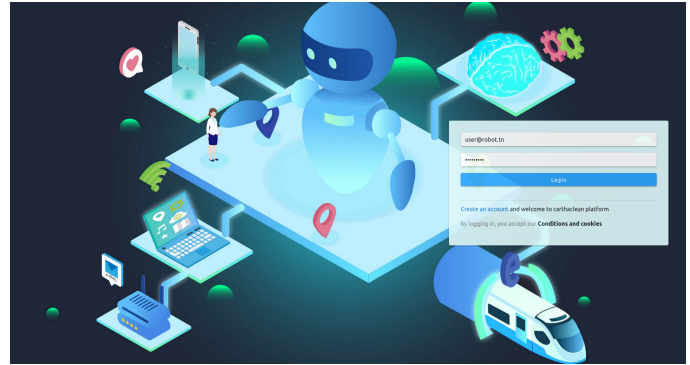

**FIG. 8** Login

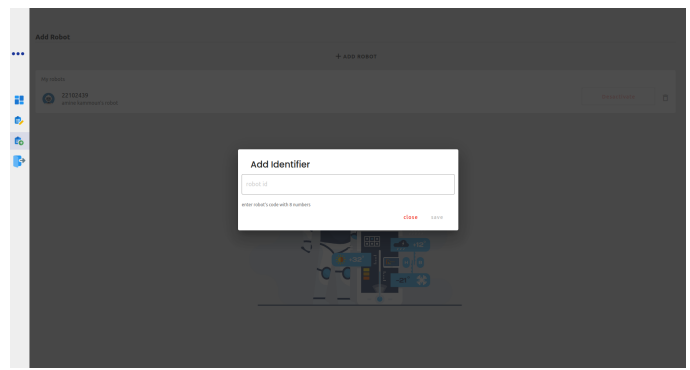

**FIG. 9** Add robot identifier

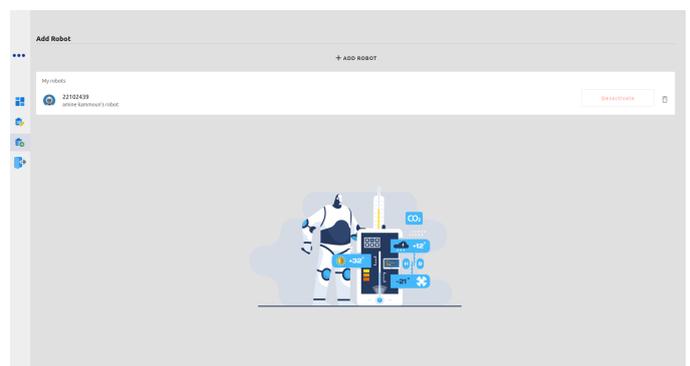

**FIG. 10** Activate your robot

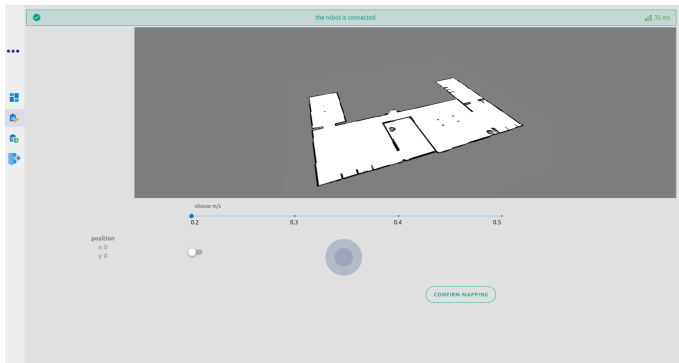

**FIG. 11** Start mapping and confirm map

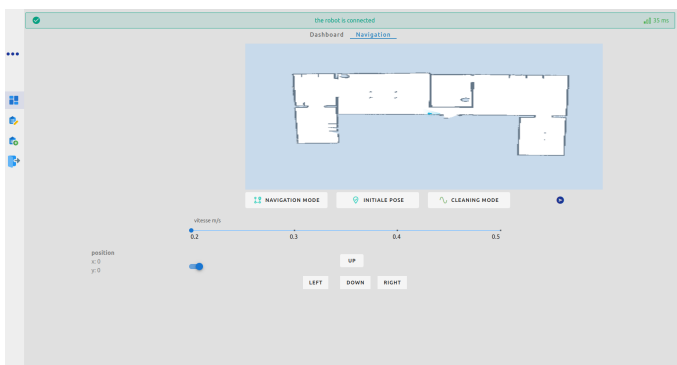

**FIG. 12** Use navigation capabilities

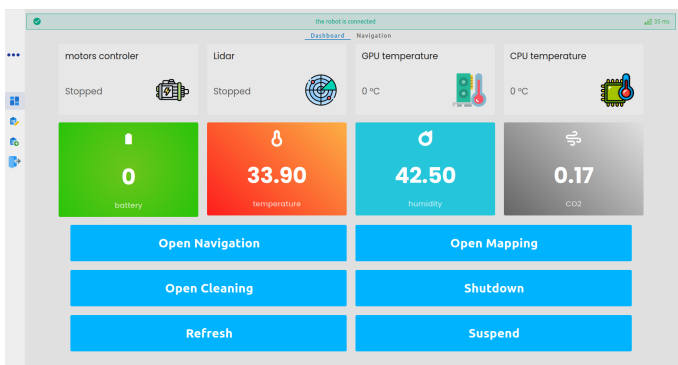

**FIG. 13** Home

## IX. WEB TO ROBOT API

The api is meant to check robot hardware status and interact with it, launching various drivers remotely and changing robot working modes.

## X. TEAM COMMITMENT & STRONG ETHICAL VALUES:

The technical support is the team who developed the robot therefore no question will go unanswered, a fast response will be delivered and a solution will be provided in no time.

We value security and privacy a lot. No data of any kind will be accessed or shared by us.
We will try to mitigate all security risks, if occured, a full recheck will be done and a security patch will be delivered. All keys and passwords will be revoked and regenerated.

## XI. CONCLUSION

We are witnessing the dawn of the robotics industry, but robots are not being created with security as a concern, often an indicator of a technology that still needs to mature. Security in robotics is often mistaken for safety. From industrial to consumer robots, going through professional ones. It's the time for all discipline to get involved in robotics developpement to design and manufacture a robot :
**Affordable** | **Safe** | **fast to Manufacture** | **Simple size and shape**| **Scalabile** | **Security**

## XII. LIMITATIONS

*With the mobility of the robot, we use batteries for power, if we take encryption to another level we will need a more power hungry CPU and GPU , and by and by our battery will be discharged faster. Every element of our system is held down by other elements.*

## XIII. FUTURE WORK

*Using qubes with ros on it, as the main operating system and use state of the art encryption.*

*Offensive Defense(Example:Honeypot)*

## XIV. REFERENCE

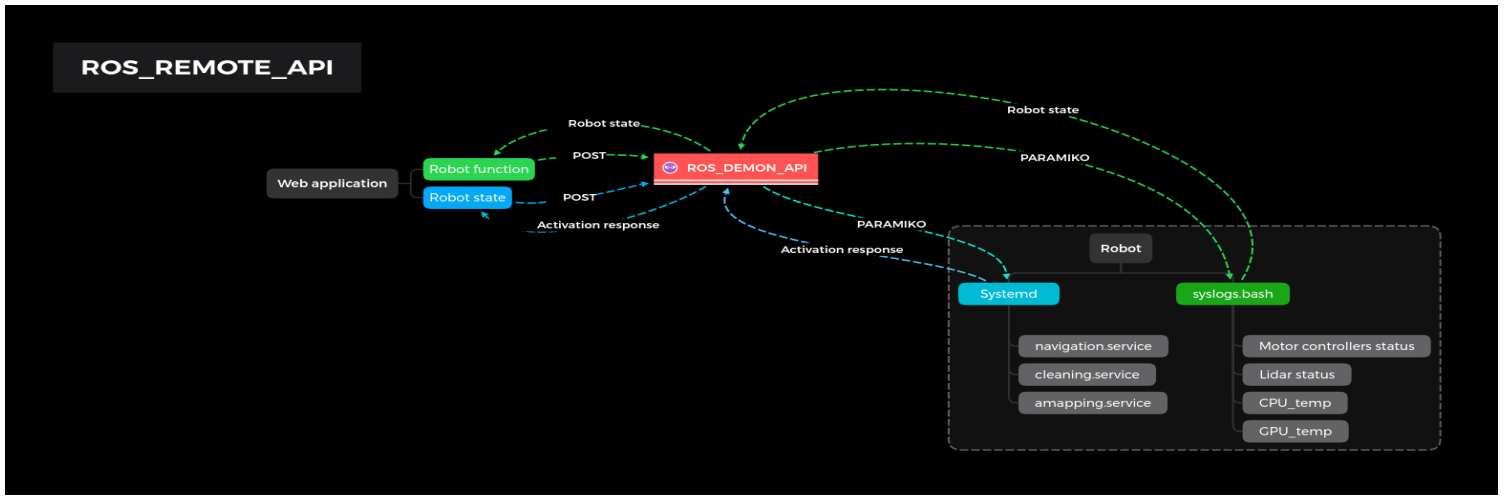